\documentclass{article}

    \PassOptionsToPackage{numbers, compress}{natbib}

    \usepackage[preprint]{neurips_2024}

\usepackage[utf8]{inputenc} 
\usepackage[T1]{fontenc}    
\usepackage{hyperref}       
\usepackage{url}            
\usepackage{booktabs}       
\usepackage{amsfonts}       
\usepackage{nicefrac}       
\usepackage{microtype}      
\usepackage{xcolor}         

\usepackage{soul}
\usepackage{graphicx}
\usepackage{amsmath,amssymb,amsfonts}
\usepackage{algorithmic}
\usepackage{caption}
\usepackage{makecell}
\usepackage{algorithmic}
\usepackage[ruled]{algorithm2e} 
\usepackage{booktabs}
\usepackage{subcaption}
\usepackage{arydshln}
\usepackage{multicol}
\usepackage{multirow}
\usepackage{tcolorbox}
\usepackage{enumitem}

\title{AdpQ: A Zero-shot Calibration Free Adaptive Post Training Quantization Method for LLMs}

\author{%
  Alireza Ghaffari$^{1,2}$ \quad Sharareh Younesian$^1$ \quad Vahid Partovi Nia$^1$ \quad\\ \textbf{Boxing Chen}$^1$ \quad \textbf{Masoud Asgharian}$^2$\\
  $^1$ Huawei Noah's Ark Lab, Montreal Research Center\\
  $^2$ Department of Mathematics and Statistics, McGill University\\
  \texttt{alireza.ghaffari@mcgill.ca} \\
  }

\begin{document}

\maketitle

\begin{abstract}
The ever-growing computational complexity of Large Language Models (LLMs) necessitates efficient deployment strategies. The current state-of-the-art approaches for Post-training Quantization (PTQ) often require calibration to achieve the desired accuracy. 
This paper presents \textbf{AdpQ}, a novel \textit{zero-shot} adaptive PTQ method for LLMs that achieves the state-of-the-art performance in low-precision quantization (e.g. 3-bit) without requiring any calibration data.
Inspired by \textit{Adaptive LASSO} regression model, our proposed approach tackles the challenge of outlier activations by separating salient weights using an \textit{adaptive soft-thresholding} method. Guided by Adaptive LASSO, this method ensures that the quantized weights distribution closely follows the originally trained weights and eliminates the need for calibration data entirely, setting our method apart from popular approaches such as SpQR and AWQ. 
Furthermore, our method offers an additional benefit in terms of privacy preservation by eliminating any calibration or training data.
We also delve deeper into the information-theoretic underpinnings of the proposed method. We demonstrate that it leverages the Adaptive LASSO to minimize the Kullback-Leibler divergence between the quantized weights and the originally trained weights. This minimization ensures the quantized model retains the Shannon information content of the original model to a great extent, guaranteeing efficient deployment without sacrificing accuracy or information. Our results achieve the same accuracy as the existing methods on various LLM benchmarks while the quantization time is reduced by at least 10x, solidifying our contribution to efficient and privacy-preserving LLM deployment.

\end{abstract}

\section{Introduction}
The rise of Large Language Models (LLMs) has fundamentally reshaped our society by performing various tasks that demand sophisticated language processing capabilities. However, LLMs versatility comes at the cost of high power consumption and memory usage. 
Since training or fine-tuning LLMs are compute-intensive and costly, Post-training Quantization (PTQ) techniques have emerged as a promising solution to address these challenges without requiring additional training or fine-tuning.
Yet, most existing PTQ algorithms rely on calibrating the weights of the LLMs using a calibration dataset, introducing additional computational overhead, time consumption, and potential data privacy concerns. Additionally, many PTQ methods are not truly zero-shot, meaning that PTQ calibration strategies can be seen as a fine-tuning step, even though the calibration does not use the same loss function as the fine-tuning stage. 

We propose \textbf{AdpQ}, a novel, zero-shot, calibration free, and adaptive PTQ method that is designed for LLMs. In our proposed approach, we leverage the power of shrinkage methods \cite{copas1983regression}, specifically the Adaptive LASSO \cite{zou2006adaptiveLASSO}, to effectively quantize LLM weights without performing any calibration. Traditionally, shrinkage methods are vastly used in statistical machine learning for model selection and variable selection. Here, we use Adaptive LASSO  to identify and isolate the weights that are more salient. Furthermore, our method achieves complete weight quantization by applying separate scales to both outlier and non-outlier weights during the quantization process, eliminating all floating-point weights in the quantized model. As will be discussed in detail throughout this paper, the main advantage of Adaptive LASSO is that it penalizes and isolates weights proportional to their originally trained values leading to an effective, calibration free outlier detection. While most popular PTQ methods use a calibration dataset to identify and tune the weights by considering activations \cite{optq,dettmers2024spqr,awq,huang2024billm}, we argue that the information of activations is somehow encoded in the weights of the model and our proposed approach helps to preserve that information considerably.

The theoretical foundation of our method is rooted in information theory. We study the Adaptive LASSO through the lens of Shannon entropy and demonstrate that Adaptive LASSO is a way to minimize the Kullback-Leibler (KL) divergence between the quantized weights and the originally trained weights leading to a more effective quantization scheme.

Our proposed PTQ method is faster than the state-of-the-art in terms of the time required to perform the quantization. We show that the Adaptive LASSO leads to a computationally efficient soft-thresholding approach, significantly reducing the run-time of the quantization algorithm. Furthermore, our method leverages mixed-precision quantization, allowing for separate quantization bit-width of outlier weights and non-outlier weights that can potentially reduce the quantization error mitigating the potential accuracy drops typically seen in low-precision quantization.

To summarize, we make the following contributions
\begin{itemize}
    \item We proposed AdpQ,  a zero-shot, calibration free, and adaptive PTQ method for LLMs that is inspired by Adaptive LASSO regression. The proposed method only relies on the weights of the model and uses \textit{no external data} for calibration while achieving the state-of-the-art accuracy performance with \textit{10$\times$ faster} quantization speed. To the best of our knowledge, this is the first time a zero-shot, calibration free PTQ is proposed that is based on Adaptive LASSO shrinkage method.

    \item The proposed method is computationally efficient and reduces the cost and run-time of the PTQ algorithm. More specifically, our proposed method can be simplified to an adaptive soft-thresholding algorithm which is easy to implement on any general-purpose hardware. 

    \item Our proposed method applies separate scales to outlier and non-outlier weights during the quantization, eliminating all floating-point weight representations in the model. Note that other state-of-the-art methods keep outliers weights in floating-point format.

    \item We provide a theoretical foundation for our proposed PTQ method using information theory. We show that our adaptive quantization method tries to minimize a quadratic error loss using quantized weights while reasonably bounding the KL-divergence between the quantized weights and the originally trained weights, hence preserving the Shannon information contents of the original weights. To the best of our knowledge, this is the first time that a quantization method, has been studied from an information-theoretic perspective.
\end{itemize}

The rest of the paper is organized as follows. Section \ref{sec:related-works} reviews recent works in the field of PTQ and specifies the differences to our proposed adaptive quantization method. Section \ref{sec:method} discusses the AdpQ algorithm in detail. Section \ref{sec:theory} delves deeper into the theoretical analysis of the algorithm and shows how AdpQ can control information loss in quantization. 
Finally, experimental results supporting our proposed methodology and theoretical findings are presented in Section \ref{sec:experiment}.

\section{Related Works}\label{sec:related-works}

In the field of low-precision deep learning, three existing notable categories are (i) low-precision or quantized training, (ii) quantization-aware training (QAT), and (iii) post-training quantization (PTQ). While our proposed method can be applied to both low-precision training (e.g. \citep{banner2018scalable,zhang2020fixed,zhu2020towards,zhao2021distribution,ghaffari2022integer}) and QAT (e.g. \cite{zhu2023survey,dettmers2024qlora,liu2023llm} ), the primary focus of this section is to review the research around PTQ of LLMs which is found to be more challenging in the literature.

Historically, PTQ methods were common for computer vision models with small number of parameters, some notable methods are AdaRound \cite{adaround}, OBQ \cite{obc}, AdaQuant \cite{adaquant}, and BRECQ \cite{brecq}. However, these methods were found to be either compute-intensive or inaccurate for large language models.

LLM.int8() \cite{llm-int8} and ZeroQuant \cite{zeroquant} are among the first PTQ techniques that were designed for LLMs. LLM.int8() separates the outlier activations and keeps them in floating-point number format while quantizing weights and non-outlier activations to 8-bit integers. LLM.int8() separates the outlier activations based on their magnitude. On the other hand, ZeroQuant uses a fine-grained hardware-friendly quantization scheme as well as layer-by-layer knowledge distillation for quantizing both weight and activations. However, both LLM.int8() and ZeroQuant are not efficient for quantizing LLMs to extreme low-percision number formats such as 3-bit integers.
OPTQ \cite{optq} is a PTQ algorithm for LLMs that can quantize weights to 3- or 4-bit integers. OPTQ adapted a calibration algorithm inspired by \cite{obs} that minimizes the $\ell_2$ loss of the quantized layer output with the original output. SpQR \cite{dettmers2024spqr} uses OPTQ algorithm while separating the salient weights and keeping them in FP16 format and further uses double quantization to reduce the memory. Both SpQR and OPTQ algorithms require calibration data for quantization.
SmoothQuant \cite{smoothquant} performs 8-bit integer quantization of weights and activation by offline migration of the quantization difficulty from activations to weights. Likewise, AWQ \cite{awq}, quantized weights by applying per-channel scales that protect the salient weights by observing the activation. SmoothQuant and AWQ algorithms also require calibration data to perform quantization. 

The key feature of our proposed algorithm, AdpQ, lies in its ability to perform PTQ without requiring any calibration data as opposed to OPTQ, SpQR, AWQ, and SmoothQuant. Furthermore, AdpQ uniquely identifies and isolates outlier weights solely through analysis of the model weight tensors. Our proposed algorithm quantizes both outlier and non-outlier weights to 3- or 4-bit integer numbers, achieving a completely quantized model without any remaining floating-point weight values. Our experiments demonstrate that AdpQ delivers at least 10$\times$ faster quantization run-time compared to the state-of-the-art as mentioned above algorithms. Finally, AdpQ is designed with information theory in mind. By focusing on retaining the information encoded within the model weights, it aims to minimize information loss during the quantization process.

\section{Methodology}\label{sec:method}

Having established the limitations of existing PTQ methods and the advantages of our proposed zero-shot approach, this section provides a comprehensive understanding of the inner workings of AdpQ while its information-theoretic foundations are left to be discussed in Section \ref{sec:theory}.

\subsection{Existence of the Best Proxy for Weights and Adaptive LASSO}

In Post-Training Quantization (PTQ), the most critical information lies within the model's weights. 
 Here, we draw inspiration from the concept of shrinkage in statistical machine learning.  Shrinkage techniques prioritize retaining the most important variables in the model while shrinking less influential ones towards zero. Similarly, outlier weight identification in PTQ can be viewed as a form of shrinkage.
 While traditional LASSO is commonly used for variable selection in statistics, it can lead to inconsistencies in the context of PTQ.  To address this, we leverage the concept of Adaptive LASSO.  This method incorporates adaptive weights to penalize different coefficients in the $\ell_1$ penalty. This allows us to effectively identify and isolate outlier weights within the model. Consider the following optimization problem,
 \begin{equation}
    \arg\min_{\hat{\textbf{W}}} \| \textbf{W}\textbf{X} - \hat{\textbf{W}}\textbf{X} \|_2^2 + \lambda \sum_{i} \left |\frac{\hat w_i}{w_i} \right|,
    \label{eq:adaptive-lasso}
\end{equation}
where $\textbf{X}$ denotes the input tensor of a layer, ${\textbf{W}}$ is the original weights of the layer, $\hat{\textbf{W}}$ is the quantized weights, $w_i$ and $\hat w_i$ are elements of ${\text{vec}(\textbf{W})}$ and ${\text{vec}(\hat{\textbf{W}})}$ respectively, and $\lambda$ is the shrinkage factor. When $\lambda$ is large, all weights are shrunk to zero and as $\lambda$ decreases, important weights are gradually chosen to be non-zero. Ultimately, when $\lambda$ is zero,  the equation \eqref{eq:adaptive-lasso} transforms to OPTQ optimization i.e. $\arg\min_{\hat{\textbf{W}}} \| \textbf{W}\textbf{X} - \hat{\textbf{W}}\textbf{X} \|_2^2$. We emphasize that Adaptive LASSO has a relative $\ell_1$ penalty term ensuring that the weights are shrunk relative to their original values, as opposed to LASSO which penalizes larger weights without considering their original values. 
 
\subsection{Adaptive LASSO as a Soft-Thresholding Method}
Since equation \eqref{eq:adaptive-lasso} does not have a closed-form solution, one may use an iterative numerical solution to solve it for a given $\lambda$. However, here we show that under a mild regulatory condition, Adaptive LASSO is a soft-thresholding algorithm.

Let us assume $\textbf{X}$ is an orthogonal matrix i.e. $\textbf{X}\textbf{X}^\top=b\textbf{I}$, where $b$ is a constant and $\textbf{I}$ is the identity matrix. By expanding \eqref{eq:adaptive-lasso} we have
\begin{align}
      &\mathcal{L} = \big (\textbf{W}\textbf{X} - \hat{\textbf{W}}\textbf{X} \big ) \big (\textbf{W}\textbf{X} - \hat{\textbf{W}}\textbf{X} \big )^\top + \lambda \sum_{i} \left |\frac{\hat w_i}{w_i} \right|= b\|{\textbf{W}}\|_2^2 - 2b\hat{\textbf{W}} \textbf{W}^\top + b\|\hat{\textbf{W}}\|_2^2 + \lambda \sum_{i} \left|\frac{\hat w_i}{w_i}\right|,
      \label{eq:soft-th-dev1}
\end{align}
and since $\textbf{W}$, weights of the original model are constant, the Adaptive LASSO loss becomes
 \begin{align}
      & \mathcal{L} = -2b\hat{\textbf{W}} \textbf{W}^\top + b\|\hat{\textbf{W}}\|_2^2 + \lambda \sum_{i} \left|\frac{\hat w_i}{w_i}\right| = \sum_{i} \big ( -2b w_i \hat{w}_i + b\hat{w}_i^2 + \lambda \left |\frac{\hat{w}_i}{w_i} \right| \big ).
      \label{eq:soft-th-dev3}
\end{align}
By taking the derivative with respect to $\hat{w}_i$, and setting it equal to zero, it is easy to see
 \begin{align}
      & \hat{w}_i = \text{sign}({w}_i) \text{{ReLU}}(|{w}_i|-\frac{\lambda'}{|{w}_i|}),
      \label{eq:soft-th}
\end{align}
where $\lambda' = {\lambda}/{2b}$ and $\text{{ReLU}}()$ denotes the positive part i.e. $\text{{ReLU}}(x) = \max (x,0)$. Equation \eqref{eq:soft-th} shows that Adaptive LASSO is a simple soft-thresholding method that is very efficient to be implemented in the currently available commodity hardware.

\textbf{Remark 1:} LLMs typically operate on high dimensional data,  particularly in the case of large input sequences. Besides the well-known curse of dimensionality, high dimensional data possess some intriguing properties as well, which is commonly known as \textit{blessing of dimensionality} \cite{donoho2000high}. One notable and interesting blessing of high dimensional data, known as the \textit{concentration phenomenon}, is that random vectors tend to be orthogonal as shown in \cite[equation (2)]{hall2005geometric}, and in \cite[equation (3)]{zorich2015multidimensional}. Therefore, we argue that the orthogonality assumption, $\textbf{X}\textbf{X}^\top=b\textbf{I}$, is not a restrictive assumption in LLMs.
\begin{figure*}[!h]
\hspace{-0.5cm}\begin{minipage}{.6\textwidth}
  \centering
  \hspace{-2cm}\includegraphics[width=0.999\linewidth]{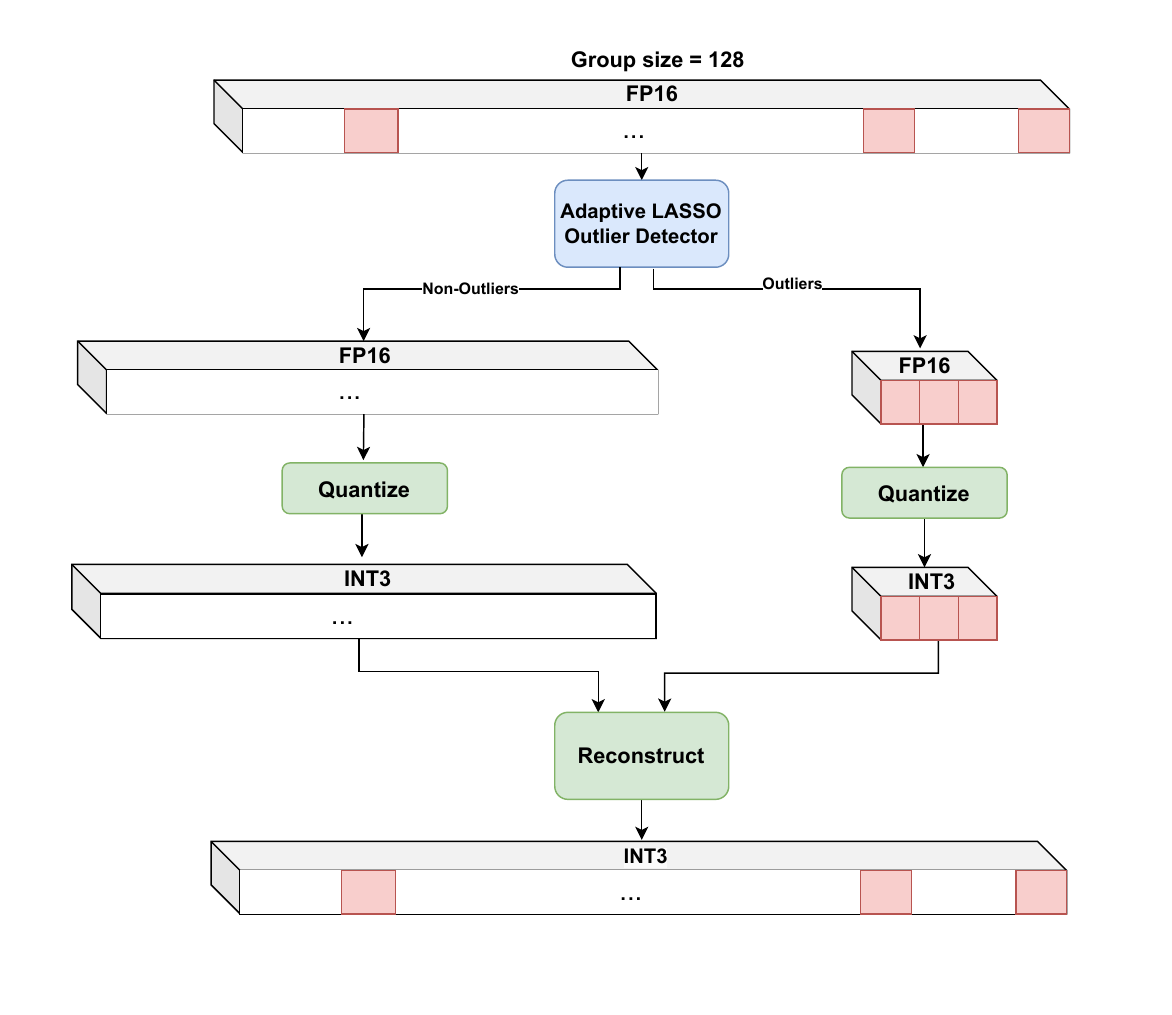} 
\caption{AdpQ outlier separation for weights.}
  \label{fig:AdpQ}
  \end{minipage}
\hspace{-1cm}\begin{minipage}{.55\linewidth}
{
\centering
\begin{algorithm}[H]{
\textbf{Input:} Layer weight tensor $\textbf{W}$, Outlier percentage $\alpha$\\
\textbf{Step 1.} Start from  a large $\lambda' >> 0$ in \eqref{eq:soft-th} then reduce it  until $\alpha$ percent of weights are selected as outlier. \\
\textbf{Step 2.} Quantize outlier weights and non-outlier weights using minmax quantization\\
\textbf{return:}  Reconstruct $\hat{\textbf{W}}$

\caption{AdpQ Quantization algorithm}
\label{alg:al_alg}
}\end{algorithm}
}
\end{minipage}
\end{figure*}%

\subsection{AdpQ Algorithm}
To provide a comprehensive overview of AdpQ algorithm, Figure \ref{fig:AdpQ} and Algorithm \ref{alg:al_alg} offer a visual and step-by-step breakdown of the procedure.
Figure \ref{fig:AdpQ} illustrates the core principle of AdpQ. It highlights how the Adaptive LASSO is used to identify and isolate outlier weights within the model. Moreover, the figure shows both outlier and non-outlier weights are quantized to low-precision formats.
Algorithm \ref{alg:al_alg} outlines the implementation process of AdpQ quantization in a clear two-step procedure. Starting with a large $\lambda'$ value (regularization parameter), the algorithm initially sets all weights to zero. As $\lambda'$ is gradually reduced, the algorithm progressively selects important weights. Once a predefined percentage, $\alpha$, of weights are identified as outliers, the process transitions to quantizing both the outlier and non-outlier weights using a minmax quantization approach.

\section{Theoretical Justification}\label{sec:theory}
Having established the methodology of AdpQ algorithm, this section delves into the information-theoretic foundation that solidifies AdpQ effectiveness. Since the distribution of the quantized weights must closely follow the distribution of the original weight to retain the information content of the model, we formulate the objective function of quantization as
 \begin{equation}
    \arg\min_{\hat{\textbf{W}}} \| \textbf{W}\textbf{X} - \hat{\textbf{W}}\textbf{X} \|_2^2 + \lambda \mathbb{D}_{\text{KL}}( f_{{\textbf{W}}}\| f_{\hat{\textbf{W}}}   ),
    \label{eq:obj_quant_KL}
\end{equation}
where $f_{\hat{\textbf{W}}}$ and $f_{{\textbf{W}}}$ are distributions of $\hat{\textbf{W}}$ and ${\textbf{W}}$ respectively and $\mathbb{D}_{\text{KL}}$ denotes the KL-divergence of the distributions. Our objective is to show that (i) Adaptive LASSO, equation \eqref{eq:adaptive-lasso}, is a proxy solution to the minimization problem \eqref{eq:obj_quant_KL} as shown in \textbf{\textit{Proposition 1}},  and (ii) separating the outlier weights improves the quantization accuracy as shown in \textbf{\textit{Proposition 2}}.

\textbf{Remark 2:} Although inputs tensor $\textbf{X}$ appears in minimization problem \eqref{eq:obj_quant_KL}, the final algorithm is simplified as a soft-thresholding method on weights tensor. Thus, our proposed PTQ algorithm does not use any calibration data.

\textbf{Proposition 1:} Suppose $f_{\textbf{W}}$ is twice continuously differentiable. Let $f'_{\textbf{W}}$ and $f''_{\textbf{W}}$ denote the first and second derivatives of $f_{\textbf{W}}$. Suppose the mean $\mu_{\delta}$ and variance $\sigma^2_{\delta}$ of the quantization error $\delta$ are small. Then
\begin{enumerate}[label=]
    \item \textbf{Claim 1:} $\mathbb{D}_{\text{KL}}( f_{{\textbf{W}}}\| f_{\hat{\textbf{W}}}) \approx \mu_{\delta} \sum_{i} f'_{{\textbf{W}}}(w_i) + \mu_{\delta} \sum_{i} f''_{{\textbf{W}}}( w_i) (\hat w_i - w_i)$
    \item \textbf{Claim 2:}   $\left |\mu_{\delta} \sum_{i} f''_{{\textbf{W}}}( w_i) (\hat w_i - w_i) \right| \leq C \left ( \sum_{i} \left |\frac{\hat w_i}{w_i} \right| + 1 \right )$ where $C$ is a constant.
\end{enumerate}


\textbf{\textit{Proof:}} Assuming a quantization error $\delta_{i}$, each original weight relates to the quantized weight such that $\hat{w}_i = {w}_i + \delta_{i}$. Let us also assume errors $\delta$ are independent of the weights values. Therefore, quantized weight distribution is a convolution of original weights distribution and quantization error distribution such that
\begin{align}
    f_{\hat{\textbf{W}}}(\hat w) &= (f_{{\textbf{W}}} \ast f_{\delta})(\hat w) = 
    \int_{-\infty}^{\infty}f_{{\textbf{W}}}(\hat w-x)f_{\delta}(x)dx\\\nonumber
    &=f_{{\textbf{W}}}(\hat w) + \int_{-\infty}^{\infty}(f_{{\textbf{W}}}(\hat w-x)-f_{{\textbf{W}}}(\hat w))f_{\delta}(x)dx.
    \label{eq:conv_dist}
\end{align}
Using the mean value theorem for $f_{{\textbf{W}}}(\hat w-x)-f_{{\textbf{W}}}(\hat w)$, we have
\begin{align}
    f_{\hat{\textbf{W}}}(\hat w) \nonumber &=f_{{\textbf{W}}}(\hat w) + \int_{-\infty}^{\infty}(-x)f'_{{\textbf{W}}}(\xi_{\hat w}(x))f_{\delta}(x)dx\\
    & \stackrel{\sigma^2_{\delta} \text{ is small}}{\approx} f_{{\textbf{W}}}(\hat w) - \int_{-\infty}^{\infty}xf'_{{\textbf{W}}}(\hat w)f_{\delta}(x)dx, \\\nonumber
\end{align}
and thus,
\begin{align}
    \frac{f_{\hat{\textbf{W}}}(\hat w)}{f_{{\textbf{W}}}(\hat w)} \approx 1 - \int_{-\infty}^{\infty}\frac{xf'_{{\textbf{W}}}(\hat w)}{f_{{\textbf{W}}}(\hat w)}f_{\delta}(x)dx
    = 1 - \frac{f'_{{\textbf{W}}}(\hat w)}{f_{{\textbf{W}}}(\hat w)}\int_{-\infty}^{\infty} xf_{\delta}(x)dx = 1 - \mu_{\delta}\frac{f'_{{\textbf{W}}}(\hat w)}{f_{{\textbf{W}}}(\hat w)},
\end{align}
where $\mu_{\delta}$ is the mean of the quantization error $\delta$. Then
\begin{align}
    \ln \left (\frac{f_{\hat{\textbf{W}}}(\hat w)}{f_{{\textbf{W}}}(\hat w)} \right ) \approx 
    \ln \left ( 1 - \mu_{\delta}\frac{f'_{{\textbf{W}}}(\hat w)}{f_{{\textbf{W}}}(\hat w)} \right ) 
    \stackrel{|\mu_{\delta}| \text{ is small}}{\approx} - \mu_{\delta}\frac{f'_{{\textbf{W}}}(\hat w)}{f_{{\textbf{W}}}(\hat w)} 
    \label{eq:dev1_kl}
\end{align}
By plugging the equation \eqref{eq:dev1_kl} in KL divergence, we have
\begin{align}
    \mathbb{D}_{\text{KL}}( f_{{\textbf{W}}}\| f_{\hat{\textbf{W}}})= -\sum_{i} f_{{\textbf{W}}}(\hat w_i) \ln \left (\frac{f_{\hat{\textbf{W}}}(\hat w_i)}{f_{{\textbf{W}}}(\hat w_i)} \right ) \approx \mu_{\delta}  \sum_{i} f'_{{\textbf{W}}}(\hat w_i).
\end{align}
Then, it follows from Taylor's expansion around the original weight $w_i$, i.e.  $f'_{{\textbf{W}}}(\hat w_i) \approx f'_{{\textbf{W}}}( w_i) + f''_{{\textbf{W}}}( w_i) (\hat w_i - w_i) $ that 
\begin{align}
    \mathbb{D}_{\text{KL}}( f_{{\textbf{W}}}\| f_{\hat{\textbf{W}}}) \approx \mu_{\delta} \sum_{i} f'_{{\textbf{W}}}(w_i) + \mu_{\delta} \sum_{i} f''_{{\textbf{W}}}( w_i) (\hat w_i - w_i),
    \label{eq:dev2_kl}
\end{align}
which proves \textbf{Claim 1}.

To prove \textbf{Claim 2}, 
since $w_i$ and $f''(w_i)$ are bounded, i.e. $|w_i| \leq A$ and $|f''(w_i)| \leq B$ in which $A$ and $B$ are constants, using triangular inequality
\begin{align}
    |\mu_{\delta} \sum_{i} f''_{{\textbf{W}}}( w_i) (\hat w_i - w_i)| = |\mu_{\delta}| \sum_{i} |w_i| |f''_{{\textbf{W}}}( w_i)| \left | \left  (\frac {\hat w_i} {w_i} - 1 \right )\right | \leq C \left (\sum_{i} \left |\frac{\hat w_i}{w_i} \right | + 1 \right ),
    \label{eq:dev3_kl}
\end{align}
in which $C = |\mu_{\delta} |AB$. This completes the proof.

\textbf{Remark 3: }Since in PTQ, original weights, and their distribution are known, the first term in equation \eqref{eq:dev2_kl} is constant. Therefore, minimizing $\mathbb{D}_{\text{KL}}( f_{{\textbf{W}}}\| f_{\hat{\textbf{W}}})$ is almost like minimizing $\mu_{\delta} \sum_{i} f''_{{\textbf{W}}}( w_i) (\hat w_i - w_i)$. Thus, following inequality \eqref{eq:dev3_kl}, we may replace $\mathbb{D}_{\text{KL}}( f_{{\textbf{W}}}\| f_{\hat{\textbf{W}}})$ with  $\sum_{i} |\frac{\hat w_i}{w_i}|$ in minimization problem \eqref{eq:obj_quant_KL} which shows Adaptive LASSO is a proxy solution to minimization problem \eqref{eq:obj_quant_KL}.


In what follows, we show that separating outlier weights guided by minimizing $\mathbb{D}_{\text{KL}}( f_{{\textbf{W}}}\| f_{\hat{\textbf{W}}})$, leads to information gain in the quantized model. Recall that Shannon entropy of a random variable $X$ with distribution $f_X(x)$ is $H(f_X):= \sum_{x}f_X(x) \ln f_X(x)$. Furthermore, the cross-entropy of random variables $X$ and $Y$ is defined to be $H(f_X,f_Y):= \sum_{x}f_X(x) \ln f_Y(x)$ and $\mathbb{D}_{\text{KL}}(f_X,f_Y)=H(f_X,f_Y)-H(X)$. We also need to recall Jenson-Shannon divergence $\textit{JSD}$ for mixture distributions. Suppose a mixture distribution $f=\pi f_X + (1-\pi)f_Y$, then $\textit{JSD}(f_X,f_Y)= \pi \mathbb{D}_{\text{KL}}(f_X\|f) + (1-\pi) \mathbb{D}_{\text{KL}}(f_Y\|f) = H(f)-(\pi H(f_X) + (1-\pi)H(f_Y))$. In the context of quantization, $\textit{JSD}$ represents the information gain due to separating outlier weights from non-outlier weights.


Let us assume $f_{{\textbf{W}}}$ is a mixture distribution of outlier weights $f^\text{O}_{{\textbf{W}}}$, and non-outlier weights  $f^\text{C}_{{\textbf{W}}}$ such that  $f_{{\textbf{W}}} = \pi f^\text{C}_{{\textbf{W}}} + (1-\pi )f^\text{O}_{{\textbf{W}}}$. We also define a similar mixture distribution for quantized weights $f_{\hat{\textbf{W}}} = \pi f^\text{C}_{\hat{\textbf{W}}} + (1-\pi )f^\text{O}_{\hat{\textbf{W}}}$ \cite{alireza_mitigating}. 

\textbf{Proposition 2:} Minimizing $\mathbb{D}_{\text{KL}}( f_{{\textbf{W}}}\| f_{\hat{\textbf{W}}})$ is the same as minimizing 
$$-\textit{\textit{JSD}}(f^\text{C}_{{\textbf{W}}}, f^\text{O}_{{\textbf{W}}}) + \underbrace{ \left \{ \pi\mathbb{D}_{\text{KL}}( f^\text{C}_{{\textbf{W}}}\| f_{\hat{\textbf{W}}}) + (1-\pi)\mathbb{D}_{\text{KL}}( f^\text{O}_{{\textbf{W}}}\| f_{\hat{\textbf{W}}})  \right \} }_{\text{Jenson-Shannon centroid for quantized weights}}$$
where by minimizing the Jensen–Shannon centroid for quantized weights, we minimize the information loss for the quantized model.

\textbf{\textit{Proof:}}
\begin{align}
    \mathbb{D}_{\text{KL}}( f_{{\textbf{W}}}\| f_{\hat{\textbf{W}}}) \nonumber &= \sum_{i} f_{{\textbf{W}}}(w_i) \ln \left (f_{\hat{\textbf{W}}}(w_i) \right ) - \sum_{i} f_{{\textbf{W}}}(w_i) \ln \left (f_{{\textbf{W}}}(w_i) \right )\\
    \nonumber &=\sum_{i} (\pi f^\text{C}_{{\textbf{W}}}(w_i) + (1-\pi)f^\text{O}_{{\textbf{W}}}(w_i)) \ln \left (f_{\hat{\textbf{W}}}(w_i) \right ) - H(f_{{\textbf{W}}})\\
    \nonumber&= \pi H(f^\text{C}_{{\textbf{W}}},f_{\hat{\textbf{W}}}) + (1-\pi) H(f^\text{O}_{{\textbf{W}}},f_{\hat{\textbf{W}}}) -H(f_{{\textbf{W}}}) \\
    \nonumber&= \pi \{H(f_{\hat{\textbf{W}}}) + \mathbb{D}_{\text{KL}}( f^\text{C}_{{\textbf{W}}}\| f_{\hat{\textbf{W}}}) \} 
    + (1-\pi) \{H(f^\text{O}_{{\textbf{W}}}) + \mathbb{D}_{\text{KL}}( f^\text{O}_{{\textbf{W}}}\| f_{\hat{\textbf{W}}} \} 
    -H(f_{{\textbf{W}}})\\
    \nonumber&= -\{H(f_{{\textbf{W}}}) - \pi H(f_{\hat{\textbf{W}}}) - (1-\pi)H(f^\text{O}_{{\textbf{W}}}) \} 
    + \pi \mathbb{D}_{\text{KL}}( f^\text{C}_{{\textbf{W}}}\| f_{\hat{\textbf{W}}}) 
    + (1-\pi) \mathbb{D}_{\text{KL}}( f^\text{O}_{{\textbf{W}}}\| f_{\hat{\textbf{W}}})\\
    &= -\textit{JSD} (f^\text{C}_{{\textbf{W}}}, f^\text{O}_{{\textbf{W}}})
    + \left \{ \pi \mathbb{D}_{\text{KL}}( f^\text{C}_{{\textbf{W}}}\| f_{\hat{\textbf{W}}}) 
    + (1-\pi) \mathbb{D}_{\text{KL}}( f^\text{O}_{{\textbf{W}}}\| f_{\hat{\textbf{W}}}) \right \},
    \label{eq:kl_information}
\end{align}
where equation \eqref{eq:kl_information}  completes the proof.

\textbf{Remark 4:} Note that the first term in equation \eqref{eq:kl_information}, i.e. $-\textit{JSD} (f^\text{C}_{{\textbf{W}}}, f^\text{O}_{{\textbf{W}}})$ is \textit{intrinsic} to the originally trained weight data and does not contribute to the minimization of $\mathbb{D}_{\text{KL}}( f_{{\textbf{W}}}\| f_{\hat{\textbf{W}}})$ in the post-training quantization. Moreover,  $-\textit{JSD} (f^\text{C}_{{\textbf{W}}}, f^\text{O}_{{\textbf{W}}})$ can be interpreted as the information loss should we decide \textit{not} to  separate the outlier weights of the trained model.

\textbf{Remark 5:} It follows from \textbf{\textit{Proposition 2}} that minimizing $\mathbb{D}_{\text{KL}}( f_{{\textbf{W}}}\| f_{\hat{\textbf{W}}})$ minimizes the Jensen-Shannon centroid for quantized weights, i.e. $\pi\mathbb{D}_{\text{KL}}( f^\text{C}_{{\textbf{W}}}\| f_{\hat{\textbf{W}}}) + (1-\pi)\mathbb{D}_{\text{KL}}( f^\text{O}_{{\textbf{W}}}\| f_{\hat{\textbf{W}}})$ leading to minimization of information loss in the quantized model. Thus, our proposed method AdpQ, as a proxy solution to optimization problem \ref{eq:obj_quant_KL}, helps retaining the information in the quantized model to a great extent.

\section{Experimental Results}\label{sec:experiment}

This section provides experimental results supporting our proposed methodology for post-training quantization of LLMs. Note that in our results,  we use a row-wise group quantization technique in conjunction with AdpQ soft-thresholding method as explained in Algorithm \ref{alg:al_alg}.





\textbf{Average bits:} The average bit presented in the results of AdpQ is calculated based on three factors, (i) the number of non-outlier weights and their bit-width, (ii) the number of outlier weights and their bitwidth and (iii) location index of the outlier weights.
Since AdpQ identifies and isolates the outlier weights in an unstructured manner, tracking the location index of the outliers is essential for tensor reconstruction after dealing with quantized outlier weights and non-outlier weights separately.
While maintaining an outlier mask is straightforward, it would add an extra bit per weight, which is inefficient in terms of memory consumption. 
To tackle this issue, we chose to retain the location index of outliers within each group when using group quantization.
Retaining the index of outlier weights leads to a lower average bit since in AdpQ, the outlier weight ratio $\alpha$, is at most 10\%. This approach results in fewer bits compared to using a mask, i.e. it requires $\log_2 g$ bits only for each outlier where $g$ is the group size. Moreover, we store scales and zero-points in 16-bit floating-point format. In summary, the average number of bits per weight is computed as
\begin{equation}
    b_{\text{avg}}=\left (b_{\text{C}}+\dfrac{2\times 16}{g} \right)\times (1-\alpha )+ \left (b_{\text{O}}+\log_2 g +\dfrac{2\times 16}{g} \right)\times \alpha
\end{equation}
where $g$ is the group size, $\alpha$ is the percentage of outlier weights, and $b_{\text{O}}$ and $b_{\text{C}}$ are the bit-widths of outlier and non-outlier weights respectively.

\textbf{Clipping Non-outlier Weights:}
We also used clipping to further reduce the $b_{\text{avg}}$ while maintaining the accuracy in our 3-bit results. The clipping is done because in 3-bit quantization, maintaining quantization accuracy requires a higher ratio of outliers. On the other hand, increasing the outlier ratio would increase $b_{\text{avg}}$ due to index tracking of outliers. We observed that applying a clipping range of 90-95\% to non-outliers yields similar accuracy compared to increasing the outlier ratio. This confirms that AdpQ can also be combined with other known quantization techniques to achieve better results.


\subsection{Experimental Results}
\textbf{Quantization Time:} Benefiting from our simple soft-thresholding technique, AdpQ eliminates the need for calibration data and weight update routines. This significantly reduces the quantization time compared to existing methods. AdpQ achieves at least 10$\times$ faster quantization speed than AWQ \cite{awq} and surpasses SpQR \cite{dettmers2024spqr} quantization time by a factor of 100$\times$ as shown in Table~\ref{tab:time}. 
\begin{table}[!h]
    \centering
  \caption{Quantization time comparison}
  \label{tab:time}
  \scalebox{0.8}{\begin{tabular}{ccccc}
    \toprule
    Model     & Method  & Avg Bits &  Quantization Time (s) $\downarrow$ \\
    \midrule
                 & AWQ (g128) & 4.25  & 838 \\
    LLaMA-7B     & SpQR & 4.63 &  10901  \\
                 & AdpQ (g128, $\alpha$=8\%)  & 4.81 &  \textbf{57} \\
    \midrule
                 & AWQ (g128) & 4.25  & 1608   \\
    LLaMA-13B    & SpQR & 4.63 & 20502   \\
                 & AdpQ (g128, $\alpha$=6\%)  & 4.67 & \textbf{116}  \\
    \midrule
                 & AWQ (g128) &  4.25 &  3740   \\
     LLaMA-30B   & SpQR &  4.63 &  24069   \\
                 & AdpQ (g128, $\alpha$=5\%) & 4.60 & \textbf{470}  \\
    \bottomrule
  \end{tabular}}
  
\end{table}
%

\textbf{Coding Ability:} AdpQ is a more robust PTQ approach since it only works with weights and does not depend on calibration data. To showcase the robustness of AdpQ, we evaluated the coding performance of quantized Code-Llama model \cite{roziere2023code} on HumanEval \cite{chen2021codex} and MBPP \cite{austin2021program} datasets. HumanEval includes 164 human handwritten programming problems with a function signature, docstring, body, and several unit tests, and MBPP consists of around 1,000 crowd-sourced Python programming problems. 
Table \ref{tab:code} shows AdpQ outperforms SpQR\cite{dettmers2024spqr}, demonstrating that if calibration data does not have the same nature as the task, using calibration data decreases the performance,  while AdpQ, is robust to such issues.
\begin{table}[!h]
    \centering
  \caption{Comparison of AdpQ results for Code-Llama models on HumanEval \cite{chen2021codex} and MBPP \cite{austin2021program}. }
  \label{tab:code}
  \scalebox{0.75}{\begin{tabular}{ccccccc}
    \toprule
    Model     & Method & Avg Bits & \multicolumn{2}{c}{Human Eval} & \multicolumn{2}{c}{MBPP} \\
    \cmidrule(r){4-5} \cmidrule(r){6-7}
     & & & pass@1 & pass@10 & pass@1 & pass@10  \\
    \midrule
                & FP16  & 16.00 & 29.63 & 59.84 & 25.87 & 63.52  \\
                & RTN (g128)  & 4.25 & 30.13 & 57.97 & 28.26 & 62.42 \\
     Code-Llama-7B   & SpQR$^*$ & 4.63 & 29.94 & 57.40 & 27.59 & 61.78  \\
                & AdpQ (g128, $\alpha$=5\%) & 4.60 & \textbf{30.34} & \textbf{58.60} & 28.03 & \textbf{62.55} \\
    \midrule
               & FP16 &  16.00 & 34.79 & 66.50 & 30.17 & 67.51  \\
                & RTN (g128) & 4.25 & 33.70 & 65.88 & 29.63 & 66.00   \\
     Code-Llama-13B   & SpQR$^*$ & 4.63 & 34.19 & 65.69 & 29.74 & 66.20   \\
                & AdpQ (g128, $\alpha$=6\%) & 4.67 & \textbf{34.79} & \textbf{66.02} & \textbf{31.36} & \textbf{66.82} \\
    \bottomrule
  \end{tabular}
  } 
  
\small{$^*$ Refer to Appendix \ref{app:hp_config} for quantization settings.}
\end{table}

\textbf{Zero-Shot Task Evaluation:} We also evaluated the accuracy of LLaMA 1 \cite{touvron2023llama} and  LLaMA 2 \cite{touvron2023llama2} models on 5 zero-shot common-sense reasoning tasks including ARC(easy and challenge) \cite{clark2018think}, HellaSwag \cite{zellers2019hellaswag}, WinoGrande \cite{sakaguchi2021winogrande} and PIQA \cite{bisk2020piqa} using LM Evaluation Harness \cite{gao2021framework}. As shown in Table \ref{tab:lmeval}, AdpQ outperforms SpQR \cite{dettmers2024spqr} in both 4-bit and 3-bit quantization, showing that zero-shot quantization coupled with Adaptive LASSO outlier detection is enough for quantization and there is no need for complex calibration-based methods.

\begin{table}[!h]
\vspace{-0.5 cm}
  \centering
  \caption{Comparison of AdpQ results on zero-shot tasks using LM Evaluation Harness \cite{gao2021framework}.}
  \label{tab:lmeval}
  \scalebox{0.78}{
  \begin{tabular}{cccccccccc}
    \toprule
    Model     && Method & Avg Bit & ARC-c & ARC-e & HellaSwag   & Winogrande & PIQA  & Avg\\
    \midrule
    \multirow{6}{*}{\rotatebox{90}{\textbf{LLaMA-7B}}}
                && FP16 &16 & 41.89 & 75.25 & 56.95 & 69.93 & 78.67 & 64.54 \\
                && RTN (g128) & 4.25 & 42.92 & 74.54 & 56.29 & 70.01 & 78.18 & 64.39  \\
                \cdashline{2-8}
                &\multirow{2}{*}{4-bit} & SpQR$^*$ & 4.63& 41.72 & 75.21 & 56.65 & 69.61 & 79.05 & 64.45 \\
                && AdpQ (g128, $\alpha=8\%$)& 4.81 & \textbf{42.15} & \textbf{75.34} & \textbf{56.72} & \textbf{70.17} & 78.56 & \textbf{64.59} \\
                \cdashline{2-8}
                &\multirow{2}{*}{3-bit} & SpQR$^*$ & 3.96 & 41.55 & 74.28 & 56.31 & 68.90 & 77.80 & 63.77 \\
                && AdpQ (g128, $\alpha=9\%$, $b_\text{O}$=4)& 3.97 & \textbf{42.32} & \textbf{74.66} & 55.94 & \textbf{69.85} & \textbf{78.40} & \textbf{64.23} \\
    \midrule
    \multirow{6}{*}{\rotatebox{90}{\textbf{LLaMA-13B }}}
               && FP16 &16 & 46.42 & 77.36 & 59.88 & 72.69 & 79.16 &  67.19 \\
                && RTN (g128)  & 4.25 & 45.82 & 76.77 & 59.37 & 72.45 & 79.71 & 66.82  \\
                \cdashline{2-8}
       &\multirow{2}{*}{4-bit} & SpQR$^*$ & 4.63  & 45.73 & 76.85 & 59.70 & 73.09 & 79.22   & 66.92 \\
                && AdpQ (g128, $\alpha=6\%$) & 4.67 & \textbf{45.99} & 76.85 & 59.41 & 73.01 & 78.94 & 66.84 \\
                \cdashline{2-8}
                &\multirow{2}{*}{3-bit} & SpQR$^*$ & 3.97 & 44.62 & 77.06 & 59.13 & 72.06 & 79.72 & 66.52  \\
                && AdpQ (g128, $\alpha=10\%$) & 3.95 & \textbf{46.25} & \textbf{77.40} & 58.56 & \textbf{73.09} & 78.78 & \textbf{66.82}\\
    \midrule
    \multirow{6}{*}{\rotatebox{90}{\textbf{LLaMA-30B }}}
                && FP16 &16  & 52.90 & 80.43 & 63.37 & 75.85 & 81.12 & 70.73  \\
                && RTN (g128)& 4.25& 52.05 & 80.77 & 62.89 & 74.19 & 80.58  & 70.10  \\
                \cdashline{2-8}
                &\multirow{2}{*}{4-bit} & SpQR$^*$ & 4.63 & 51.45 & 80.47 & 63.08 & 74.74 & 80.74 & 70.10    \\
                && AdpQ (g128, $\alpha=5\%$)& 4.60 & \textbf{51.88} & \textbf{80.77} & 63.07 & 74.19 & \textbf{80.74}  & \textbf{70.13}  \\
                \cdashline{2-8}
                &\multirow{2}{*}{3-bit} & SpQR$^*$ & 3.89 & 50.77 & 80.26 & 62.79 & 74.59 & 80.47   & 69.78  \\
                && AdpQ (g128, $\alpha=8\%$, $b_\text{O}$=4) & 3.89 & \textbf{50.94} & 80.13 & 62.49 & \textbf{75.22} & \textbf{80.96} & \textbf{69.95}  \\
    \midrule
    \multirow{6}{*}{\rotatebox{90}{\textbf{LLaMA-2-7B}}}
                 && FP16 &16  & 43.43 & 76.35 & 57.16 & 69.14 & 78.07 & 64.83   \\
                && RTN (g128)& 4.25& 43.09 & 76.18 & 56.90 & 68.67 & 77.48 & 64.46  \\
                \cdashline{2-8}
        &\multirow{2}{*}{4-bit} & SpQR$^*$ & 4.63  & 44.28 & 76.14 & 56.95 & 68.51 & 77.42 & 64.66  \\
                && AdpQ (g128, $\alpha=8\%$)& 4.81 & 43.17 & \textbf{76.39} & \textbf{57.12} & \textbf{69.77} & \textbf{77.97} &  \textbf{64.88}\\
                \cdashline{2-8}
                &\multirow{2}{*}{3-bit} & SpQR$^*$ & 3.96 & 42.41 & 75.08 & 56.39 & 68.67 & 77.86 & 64.08  \\
                && AdpQ (g128, $\alpha=10\%$)& 3.95 & \textbf{42.75} & \textbf{75.38} & \textbf{56.71} & \textbf{69.53} & 77.31 & \textbf{64.34} \\
    \midrule
    \multirow{6}{*}{\rotatebox{90}{\textbf{LLaMA-2-13B}}}
                && FP16 &16  & 48.46 & 79.42 & 60.05 & 72.38 & 79.11 & 67.88  \\
                && RTN  (g128)& 4.25& 48.12 & 78.83 & 59.74 & 72.69 & 78.67  & 67.61  \\
                \cdashline{2-8}
                &\multirow{2}{*}{4-bit} & SpQR$^*$ & 4.63 & 48.46 & 79.76 & 59.97 & 71.90 & 78.84  & 67.79  \\
                && AdpQ (g128, $\alpha=6\%$)& 4.67 & 48.38 & 79.63 & 59.89 & \textbf{72.53} & \textbf{78.94} & \textbf{67.87}  \\
                \cdashline{2-8}
                &\multirow{2}{*}{3-bit} & SpQR$^*$ & 3.97 & 46.33 & 78.16 & 59.54 & 72.45 & 78.24  & 66.94  \\
                && AdpQ (g128, $\alpha=10\%$) &3.95 & 46.08 & \textbf{78.28} & 59.20 & 71.90 & \textbf{78.51} & 66.79 \\
    \bottomrule
  \end{tabular}
  }

\small{$^*$ Refer to Appendix \ref{app:hp_config} for quantization settings.}
\end{table}

\textbf{Perplexity:} We evaluated perplexity of quantized LLaMA models on WikiText2 \cite{merity2016pointer} and C4 \cite{raffel2020exploring} datasets. Table \ref{tab:perp} shows the results comparing perplexity scores for FP16, Round to Nearest (RTN), AWQ\cite{awq}, SpQR,\cite{dettmers2024spqr} and AdpQ.
AdpQ outperforms AWQ and RTN consistently in terms of perplexity scores. Furthermore, AdpQ exhibits perplexity scores that closely follow those of SpQR, particularly for larger models. These results highlight AdpQ ability to achieve competitive accuracy while offering a significant advantage in terms of quantization time efficiency and robustness. Refer to Appendix \ref{app:extra-results} for more perplexity results on Falcon \cite{falcon40b} and OPT
\cite{zhang2022opt} models.

\begin{table}[!h]
  \caption{Comparison of AdpQ perplexity results of 4-bit \& 3-bit on WikiText2 and C4.}
  \label{tab:perp}
  \centering
  \scalebox{0.7}{
  \begin{tabular}{cccccc||cccc}
  \toprule
  &&\multicolumn{4}{c}{\textbf{4-bit}}&\multicolumn{4}{c}{\textbf{3-bit}} \\
  \cmidrule(r){3-6} \cmidrule(r){7-10}\\
    Model     & Method & Quantization setting & Avg Bits & Wiki2 $\downarrow$ & C4$\downarrow$   & Quantization setting & Avg Bits & Wiki2 $\downarrow$   & C4$\downarrow$  \\
    \hline
    \multirow{5}{*}{\rotatebox{90}{\textbf{\small{LLaMA-7B}}}}
                & FP16 &- &  16.00 & 5.67 & 7.08 &-  &  16.00 & 5.67 & 7.08\\
                & RTN  & 4bit-g128  &  4.25 & 5.96 & 7.37 & 3bit-g128 & 3.25 & 7.01 & 8.62 \\
       & AWQ  &4bit-g128  & 4.25 & 5.78 & 7.21 &3bit-g128 & 3.25 & 6.35 & 7.81   \\
                & SpQR & Refer to Appendix \ref{app:hp_config}  & 4.63 & 5.73 & 7.13 &Refer to Appendix \ref{app:hp_config}   & 3.98 & 5.87 & 7.28 \\
                & AdpQ &(4bit-g128, $\alpha=8\%$)   & 4.81 & 5.78 & 7.22 &(3bit-g128, $\alpha=9\%$, $b_\text{O}$=4)  & 3.97 & 6.07 & 7.51 \\
    \hline
    \multirow{5}{*}{\rotatebox{90}{\textbf{\small{LLaMA-13B}}}}
               & FP16 &-  &  16.00 & 5.09 & 6.61 & - &   16.00 & 5.09 & 6.61  \\
                & RTN  &4bit-g128 &  4.25 & 5.25 & 6.75 &3bit-g128   & 3.25 & 5.88 & 7.50    \\
              & AWQ &4bit-g128  &4.25 & 5.18 & 6.70 &3bit-g128  & 3.25 & 5.52 & 7.07    \\
                & SpQR &Refer to Appendix \ref{app:hp_config}   & 4.63 & \textbf{5.13} & \textbf{6.64} &Refer to Appendix \ref{app:hp_config}   & 3.97 & 5.22 & 6.72   \\    
                & AdpQ & (4bit-g128, $\alpha=6\%$)  & 4.67 & \textbf{5.15} & \textbf{6.67} &(3bit-g128, $\alpha=9\%$, $b_\text{O}$=4)  & 3.97 & 5.32 & 6.81  \\
    \hline
    \multirow{5}{*}{\rotatebox{90}{\textbf{\small{LLaMA-30B}}}}
               & FP16 &-  &  16.00 & 4.10 & 5.98 &-   &  16.00 & 4.10 & 5.98   \\
                & RTN &4bit-g128  & 4.25 & 4.23 & 6.10 &3bit-g128   & 3.25 & 4.88 & 6.59   \\
                & AWQ  &4bit-g128  & 4.25 & 4.21 & 6.05 &3bit-g128  & 3.25 & 4.61 & 6.35 \\
                & SpQR &Refer to Appendix \ref{app:hp_config}  &  4.63 & \textbf{4.14} & \textbf{6.01} &Refer to Appendix \ref{app:hp_config}    & 3.90 & \textbf{4.25} & \textbf{6.08} \\
                & AdpQ &(4bit-g128, $\alpha=5\%$)   & 4.60 & \textbf{4.16} & \textbf{6.04} &(3bit-g128, $\alpha=8\%$, $b_\text{O}$=4)  & 3.89 & \textbf{4.31} & \textbf{6.15}   \\
    \hline
    \multirow{5}{*}{\rotatebox{90}{\textbf{\small{LLaMA2-7B }}}}
                 & FP16 &-  & 16.00 & 5.47 & 6.97 &-  & 16.00 & 5.47 & 6.97  \\
                 & RTN  &4bit-g128  & 4.25 & 5.72 & 7.24 &3bit-g128   & 3.25 &6.66 & 8.40   \\
                 & AWQ &4bit-g128 & 4.25 & 5.60 & 7.12 &3bit-g128  & 3.25 & 6.24 & 7.81   \\
                 & SpQR &Refer to Appendix \ref{app:hp_config}  &  4.63 & 5.52 & 7.03 &Refer to Appendix \ref{app:hp_config}   & 3.98 & 5.66 & 7.18 \\
                 & AdpQ  &(4bit-g128, $\alpha=8\%$) & 4.81 & 5.60 & 7.12 &(3bit-g128, $\alpha=9\%$, $b_\text{O}$=4)  & 3.97 & 5.83 & 7.37 \\
    \hline
    \multirow{5}{*}{\rotatebox{90}{\textbf{\small{LLaMA2-13B}}}}
                  & FP16 &-   & 16.00 & 4.88 & 6.47  &-  & 16.00 & 4.88 & 6.47  \\
                  & RTN &4bit-g128   & 4.25 & 4.98 & 6.59 &3bit-g128  & 3.25 & 5.52 & 7.18  \\
                  & AWQ  &4bit-g128  & 4.25 &4.97 & 6.56 &3bit-g128 & 3.25 & 5.32 & 6.95   \\
                  & SpQR  & Refer to Appendix \ref{app:hp_config} & 4.63 & \textbf{4.92} & \textbf{6.51}  &Refer to Appendix \ref{app:hp_config} & 3.96 & \textbf{5.01} & \textbf{6.60}   \\
                  & AdpQ  &(4bit-g128, $\alpha=6\%$) & 4.67 & \textbf{4.93} & \textbf{6.52} &(3bit-g128, $\alpha=9\%$, $b_\text{O}$=4)  & 3.97 & \textbf{5.05} & \textbf{6.66}    \\
    \bottomrule
  \end{tabular}}
\end{table}

     



\section{Conclusion}
This paper presented AdpQ, a novel zero-shot, calibration free PTQ approach that is designed specifically for LLMs. AdpQ addresses the limitations of existing PTQ methods by leveraging the Adaptive LASSO regression for outlier identification and enabling quantization of both outlier and non-outlier weights. AdpQ provides a robust and accurate quantization scheme while achieving complete model quantization with low-precision integer representations.
Our theoretical analysis demonstrates that AdpQ is aligned with information theory principles and as such ensures the information of the model is recovered in the quantized model to a great extent.
Furthermore, our experimental results show AdpQ ability to achieve competitive accuracy on standard language modeling benchmarks while surpassing existing PTQ methods in terms of quantization robustness and quantization time efficiency. Notably, AdpQ exhibits at almost a 10$\times$ speedup in quantization time compared to AWQ and a 100$\times$ speedup in quantization time compared to SpQR, highlighting its significant advantage in deployment scenarios.
Thus, AdpQ presents a compelling PTQ solution for the efficient deployment of LLMs. Its zero-shot calibration free nature, computational efficiency, and competitive accuracy make it a valuable tool for facilitating the practical application of LLMs.

\medskip
\small{
\bibliographystyle{unsrtnat}
\bibliography{neurips}
}

\appendix

\section{Experimental Settings }\label{app:settings}

\subsection{Seed Sensitivity}\label{app:seed}
Since our proposed method, AdpQ, only uses deterministic pre-trained weights of the model and performs a soft-thresholding to identify $\alpha$ percent of outlier weights, it does not exhibit any stochastic behavior during the quantization. Furthermore, we do not use any data for calibration and thus, our algorithm is robust toward randomness in data selection. We believe this is the main advantage of our proposed algorithm.

\subsection{Calibration Datasets and Parameters }
We follow the pipelines used in SpQR\footnote{See \url{https://github.com/Vahe1994/SpQR}} and AWQ\footnote{See \url{https://github.com/mit-han-lab/llm-awq}} official implementation to generate calibration datasets. Random selection of 128 samples of length 2048 form RedPajama \cite{together2023redpajama}, C4 and RefinedWeb \cite{refinedweb} is used for quantization of LLaMA 1, LLaMa 2, OPT \cite{zhang2022opt} and Falcon \cite{falcon40b} using SpQR. For AWQ experiments 128 samples from a small subset of Pile \cite{gao2020pile} dataset is used following the AWQ's implementation.

\subsection{Hyper-Parameters and Configs} \label{app:hp_config}
\textbf{RTN:} We implemented RTN quantization method based on the implementation of AWQ which supports weight reshaping for group quantization.

\textbf{AWQ:} We used AWQ's official implementation for quantizing LLaMA and OPT models.

\textbf{SpQR:} We use SpQR's official implementation for quantizing LLaMA, Code-Llama and OPT models. Table \ref{tab:spqr config} shows the hyper-parameters used for SpQR quantization.

\begin{table}[!h]
    \centering
  \caption{Quantization configuration of SpQR}
  \label{tab:spqr config}
  \scalebox{0.8}{\begin{tabular}{cccccc}
    \toprule
    Model     & Calibration & Group &  Weight & Scales \& Zeros & Outlier  \\
    & Set &Size & Bits & Bits & Threshold\\
    \midrule
    LLaMA     & RedPajama & 16 &  4 & 3& 0.2  \\
           & RedPajama & 16 &  3 & 3& 0.25-0.28  \\
    \midrule
    Code-Llama     & RedPajama & 16 &  4& 3 & 0.2  \\
    \midrule
    OPT     & C4 & 16 &  4 & 3& 0.2  \\
           \midrule
    Falcon     & RefinedWeb & 16 &  4 & 3& 0.2  \\
    \bottomrule
  \end{tabular}}
  
\end{table}

\subsection{Hardware Settings}\label{app:hw}
We perform quantization on single NVIDIA V100-32G GPU. For evaluation using LM Evaluation Harness we use 8$\times$V100-32G GPUs for 30B models.

\subsection{Extra Experiments Results}\label{app:extra-results}
Table \ref{tab:perp opt} shows the perplexity results on OPT \cite{zhang2022opt} and Falcon\cite{falcon40b} models.

\begin{table}[!h]
  \caption{Perplexity of 4-bit OPT and Falcon models on WikiText2 and C4.}
  \label{tab:perp opt}
  \centering
  \scalebox{0.8}{
  \begin{tabular}{cccccc}
  \toprule
    Model     & Method & Quantization setting & Avg Bits & Wiki2 $\downarrow$ & C4$\downarrow$   \\
    \hline
    \multirow{5}{*}{\rotatebox{0}{\textbf{\small{OPT-6.7B}}}}
                & FP16 &- &  16.00 & 10.86 & 11.74 \\
                & RTN  & 4bit-g128  &  4.25 & 11.15 & 12.31  \\
       & AWQ  &4bit-g128  & 4.25 & 10.95 & 11.86    \\
                & SpQR & Refer to Appendix \ref{app:hp_config}  & 4.63 & 10.91 & 11.78  \\
                & AdpQ &(4bit-g128, $\alpha=6\%$)   & 4.67 & \textbf{10.86} & 11.99 \\
    \hline
    \multirow{5}{*}{\rotatebox{0}{\textbf{\small{OPT-13B}}}}
               & FP16 &-  &  16.00 & 10.13 & 11.20\\
                & RTN  &4bit-g128 &  4.25 & 10.30 & 11.51    \\
              & AWQ &4bit-g128  &4.25 & 10.29 & 11.28   \\
                & SpQR &Refer to Appendix \ref{app:hp_config}   & 4.27 & \textbf{10.22} & \textbf{11.27}  \\    
                & AdpQ & (4bit-g128, $\alpha=6\%$)  & 4.67 & \textbf{10.20} & \textbf{11.31}  \\
    \hline
    \multirow{5}{*}{\rotatebox{0}{\textbf{\small{OPT-30B}}}}
               & FP16 &-  &  16.00 & 9.55 & 10.69    \\
                & RTN &4bit-g128  & 4.25 & 9.94 & 10.94   \\
                & AWQ  &4bit-g128  & 4.25 & 9.61 & 10.74  \\
                & SpQR &Refer to Appendix \ref{app:hp_config}  &  4.63 & 9.55 & 10.71  \\
                & AdpQ &(4bit-g128, $\alpha=5\%$)   & 4.60 & 9.64 & 10.79    \\
    \hline
    \multirow{5}{*}{\rotatebox{0}{\textbf{\small{  Falcon-7B}}}}
                 & FP16 &-  & 16.00 & 6.59 & 9.50  \\
                 & RTN  &4bit-g128  & 4.25 & 6.79 & 9.79   \\
                 & SpQR &Refer to Appendix \ref{app:hp_config}  &  4.44 & \textbf{6.64} & \textbf{9.58}\\
                 & AdpQ  &(4bit-g128, $\alpha=4\%$) & 4.53 & \textbf{6.69} & \textbf{9.63}  \\
    \hline
    \multirow{5}{*}{\rotatebox{0}{\textbf{\small{ Falcon-40B}}}}
                  & FP16 &-   & 16.00 & 5.23& 7.76   \\
                  & RTN &4bit-g128   & 4.25 & 5.31 & 7.88   \\
                  & SpQR  &Refer to Appendix \ref{app:hp_config} & 4.46 & \textbf{5.26} & \textbf{7.79}  \\
                  & AdpQ  &(4bit-g128, $\alpha=5\%$) & 4.60 &\textbf{5.27} & \textbf{7.81}   \\
    \bottomrule
  \end{tabular}}
\end{table}

\end{document}